\documentclass[final]{cvpr}

\usepackage{times}
\usepackage{epsfig}
\usepackage{graphicx}
\usepackage{amsmath}
\usepackage{amssymb}

\usepackage{cases}
\usepackage{paralist}

\usepackage{xcolor}
\usepackage[font=footnotesize]{caption}

\newcommand{\abc}[1]{\textcolor{black}{#1}}
\newcommand{\aabc}[1]{\textcolor{black}{#1}}
\newcommand{\aaabc}[1]{\textcolor{black}{#1}}

\newcommand{\zq}[1]{\textcolor{black}{#1}}
\newcommand{\zqq}[1]{\textcolor{black}{#1}}

\newcommand{\NOTE}[1]{\textcolor{black}{}}
\newcommand{\ZQNOTE}[1]{\textcolor{black}{}}
\newcommand{\TODO}[1]{\textcolor{black}{}}

\newcommand{\msum}{\mathrm{sum}}

\usepackage[english]{babel}

\usepackage[pagebackref=true,breaklinks=true,colorlinks,bookmarks=false]{hyperref}

\def\cvprPaperID{350} 
\def\confYear{CVPR 2021}

\begin{document}

\title{Cross-View Cross-Scene Multi-View Crowd Counting}
\author{Qi Zhang$^1$, Wei Lin$^2$, Antoni B. Chan$^1$
       \\
       $^1$ Department of Computer Science, City University of Hong Kong, Hong Kong SAR, China\\
       {\tt\small \{qzhang364-c@my., abchan@\}cityu.edu.hk}\\
       $^2$School of Computer Science and School of Artificial Intelligence,\\
       Northwestern Polytechnical University, Xi'an, Shaanxi, China.\\
       {\tt\small elonlin24@gmail.com}\\
       }

\maketitle

\thispagestyle{empty}

\vspace{-0.4cm}
\begin{abstract}
   Multi-view crowd counting has been previously proposed to utilize multi-cameras to extend the field-of-view of a single camera, capturing more people in the scene, and improve counting performance for occluded people or those in low resolution.
   \abc{However, the current multi-view paradigm trains and tests on the same single scene and camera-views, which limits its practical application.
   In this paper, we propose a cross-view cross-scene (CVCS) multi-view crowd counting paradigm, where the training and testing occur on different scenes with arbitrary camera layouts.}
   To dynamically handle the challenge of optimal view fusion under scene and camera layout change and non-correspondence noise due to camera calibration errors or erroneous features, we propose a CVCS model that attentively selects and fuses multiple views together using camera layout geometry, and a noise view regularization method to train the model to handle non-correspondence errors.
   We also generate a large synthetic multi-camera crowd counting dataset with a large number of scenes and camera views  to capture many possible variations, which avoids the difficulty of collecting and annotating such a large real dataset.
    We then test our trained CVCS model on real multi-view counting datasets, by using unsupervised domain transfer.
   The proposed CVCS model \aaabc{trained on synthetic data outperforms the same model trained only on real data, and} achieves promising performance compared to fully supervised methods that train and test on the same single scene.

\end{abstract}

\vspace{-0.5cm}

\section{Introduction}

\begin{figure}[t]
\centering

   \includegraphics[width=0.82\linewidth]{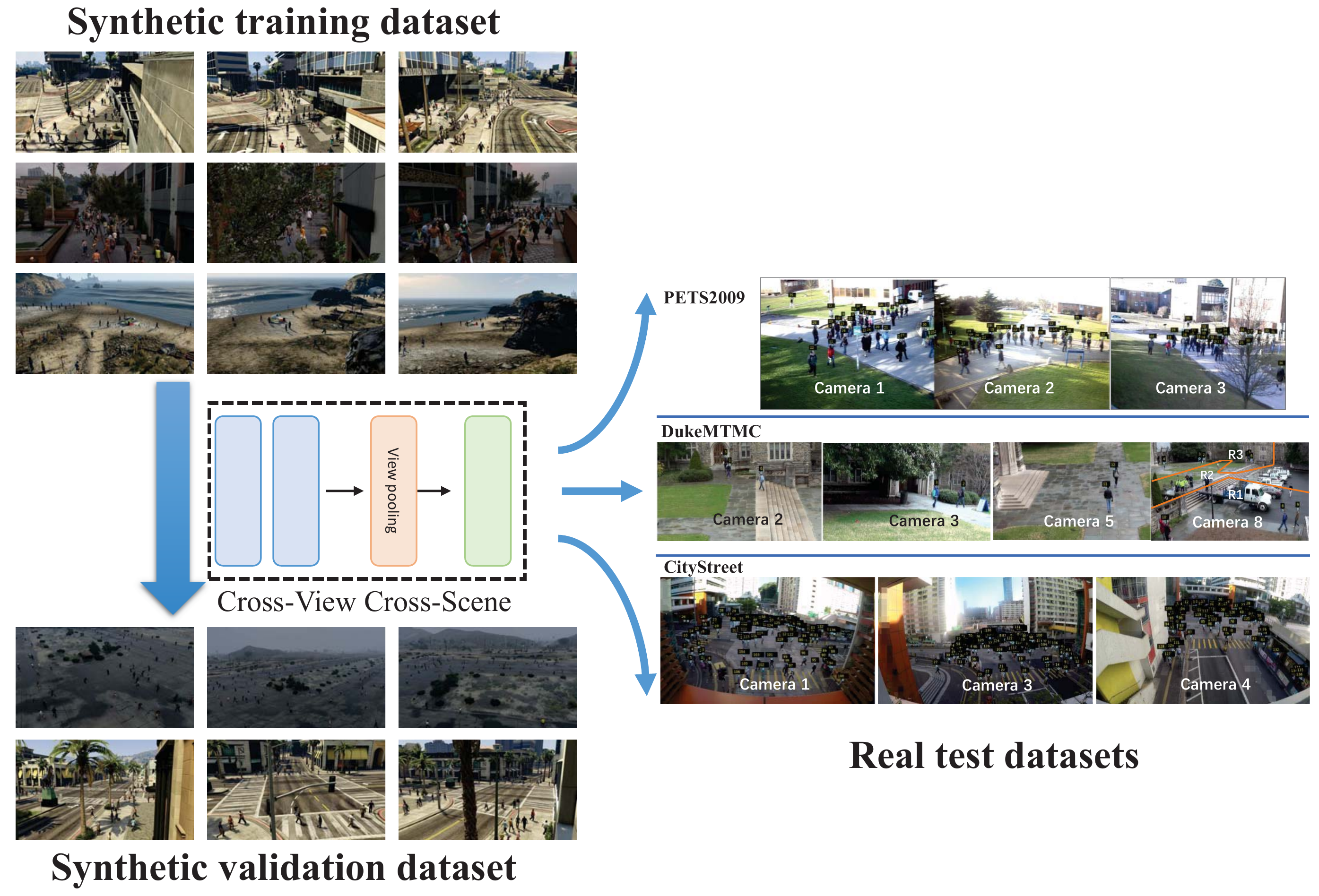}

   \vspace{-0.2cm}

   \caption{\abc{Cross-view cross-scene (CVCS) multi-view crowd counting. The CVCS model is trained and validated on synthetic multi-view crowd scenes, where the scenes and camera-views are different between the training and validation sets. To test on a real scene, unsupervised domain adaptation is applied to the trained CVCS model, where only the real images are used to fine-tune the model.}
   }
   \vspace{-0.5cm}
\label{fig:examples}
\end{figure}

\begin{figure*}[t]
\centering
   \includegraphics[width=0.82\linewidth]{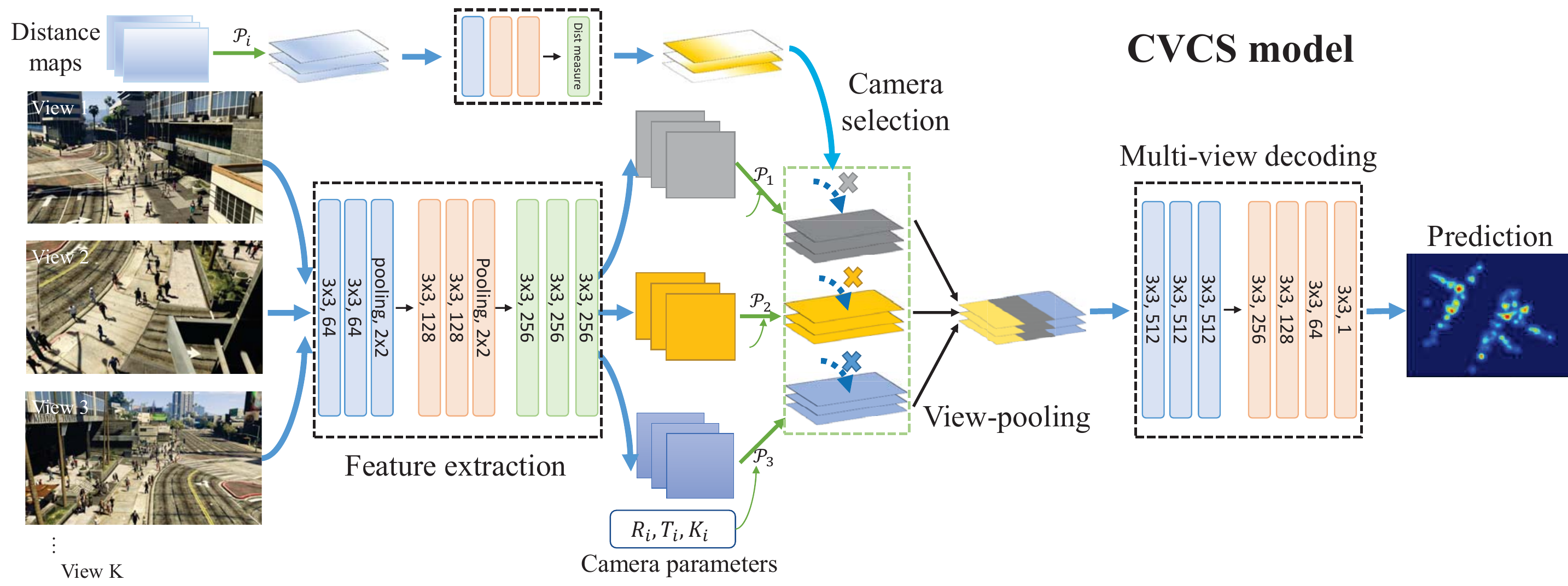}

   \vspace{-0.4cm}

   \caption{The pipeline of the cross-view cross-scene multi-view counting model (CVCS):
1) \textbf{Single-view feature extraction}: The first 7 layers of VGG-Net extracts the single-view features;
2) \textbf{Feature projection}: The extracted single-view features are projected to the average height-plane by camera projection;
3) \textbf{Multi-camera selection and fusion}: \abc{An adaptive CNN subnet selects 
 among the multi-view feature maps, through an attention mechanism guided by the object-to-camera distance}, and a max-pooling layer fuses projected camera-view features;
4) \textbf{Multi-view decoding}: The fused projected features are decoded to predict the scene-level density maps.
}
\vspace{-0.4cm}
\label{fig:main_pipeline}
\end{figure*}

\abc{\zqq{Deep neural network-based} multi-view (MV) crowd counting \cite{zhang2019wide, zhang2020_3d} was recently proposed to count people in wide-area scenes that cannot be covered by a single camera. In these works, feature maps from multiple camera views are fused together and decoded to predict a scene-level crowd density map.
However, one major disadvantage of the current MV paradigm is the models are trained and tested on the same single scene and a fixed camera layout, and thus the trained models do not generalize well to other scenes or other camera layouts.}

\abc{In this work, we propose a new paradigm of {\em cross-view cross-scene} multi-view counting (CVCS), where MV counting models are trained and tested on different scenes and arbitrary camera layouts. This paradigm is challenging because both the scene and camera layout (including number of cameras) change at test time.
In particular, in single-scene MV counting, the optimal selection of features from each camera and the handling of non-correspondence errors (caused by camera calibration errors or erroneous features) can be directly learned by the MV model (in its network parameters), since it is trained and tested on the same scene/cameras.
In contrast, for CVCS MV counting, because the camera positions, camera orders, and scenes are all varying, the MV counting model must learn to dynamically handle different camera layouts and non-correspondence noise.
To address these two issues, we propose a CVCS counting model, which attentively selects and fuses features from multiple cameras using the camera layout (object-to-camera distances), and a noise-injection regularization scheme, which simulates non-correspondence errors, improving model generalization.}


\abc{Effectively training cross-scene counting models requires a large dataset of scenes in order to capture the many possible variations of camera poses and scenes.}
For example, cross-scene models for {\em single-view} counting \cite{boominathan2016crowdnet, zhang2016single, onoro2016towards, sam2017switching, bai2020adaptive} are trained on large single-view datasets, such as ShanghaiTech \cite{zhang2016single}, UCF-QNRF \cite{idrees2018composition}, JHU-CROWD++ \cite{sindagi2019pushing} or NWPU-CROWD \cite{wang2020nwpu}, which contain thousands of images, each of a different scene.
However, current MV counting datasets, such as PETS2009 \cite{ferryman2009pets2009}, DukeMTMC \cite{ristani2016MTMC}, and CityStreet \cite{zhang2019wide} only contain 2 to 4 views of a single scene, and combining these three datasets only yields 3 scenes and 3 camera layouts, which is not enough to train a CVCS model.
Collecting and annotating a large-scale MV dataset, comprising a large number of scenes taken with many synchronized cameras, is a time-consuming and laborious task, and is further complicated due to personal privacy issues and social-distancing in the current pandemic situation.
To avoid such limitations, we generate a synthetic CVCS multi-view dataset of 31 scenes containing around 100 camera views with 100 frames in each view.  \abc{The large number of camera views for each scene is sufficient for generalizing across camera layouts.}

We use the synthetic dataset to train our CVCS model, and directly applying the trained CVCS model to real-world multi-view counting datasets, yielding promising results.
The results are further improved by using unsupervised domain adaptation to fine-tune the trained  model on only the test images (and not crowd labels).

In summary, the contributions of this paper are 3-fold:
\begin{compactenum}
  \item We propose a cross-view cross-scene multi-view counting DNN model (CVCS), which adaptively selects and fuses multi-cameras, and a noise view
        regularization method to improve generalization. To our knowledge, this is the first study on the cross-view cross-scene multi-view problem in crowd counting.
  \item We propose a large synthetic multi-view crowd counting dataset, which contains a large number of camera views, scene variations and frames. This
        is the first large synthetic dataset for multi-view counting, which enables research on cross-scene cross-view problems.
  \item The proposed CVCS model outperforms existing state-of-the-art MV models in the cross-view cross-scene paradigm.
        Furthermore, the CVCS model, trained on synthetic scenes and adapted to a real-world test scene with unsupervised domain adaptation, achieves promising performance compared with MV models trained on single-scenes.
\end{compactenum}

\vspace{-0.2cm}
\section{Related Work}
\vspace{-0.1cm}


\textbf{Multi-view crowd counting.}
Traditional multi-view (MV) counting methods \cite{li2012people, Maddalena2014people, Ryan2014Scene, Tang2014Cross, Ge2010Crowd} rely on foreground extraction techniques and hand-crafted features, 
and frequently train on PETS2009, which only contains 2 to 4 camera views and hundreds of frames.
%
%
\abc{More recently, \cite{zhang2019wide} proposed a multi-view multi-scale (MVMS) counting model, which fuses multiple views and multiple scales of feature maps into a scene-level feature map, and then decodes it to predict a scene-level density map.}
 \cite{zhang2019wide} collected a new MV counting dataset CityStreet for large-crowd single-scene training.
Follow-up work \cite{zhang2020_3d} 
proposed to use 3D ground-truth for MV counting to improve counting performance.
\abc{However, both these works are trained and tested on single scenes, and do not generalize to cross-scene tasks.}
Furthermore, the existing MV counting datasets have too few views for cross-view training.
\abc{In contrast, in our paper we address learning multi-view models that generalize well to new scenes and camera-view layouts, and generate a new large-scale synthetic dataset for training.}


\textbf{Cross-scene single-view crowd counting.}
Single-view counting algorithms \cite{Wan2019AdaptiveDM} achieve cross-scene ability mainly by training DNNs on single-view counting datasets, which contain a large number of images capturing different scenes from different view angles \cite{zhang2015cross,zhang2016single,idrees2018composition,sindagi2019pushing, wang2020nwpu, 2020NAS}.
Unsupervised/semi-supervised \cite{sam2019almost, sindagi2020learning, liu2020semi} or weakly supervised methods \cite{liu2019exploiting, borstel2016gaussian, yangweakly,zhao2020active} have also been proposed to improve the cross-scene counting performance on the existing single-view datasets.
Various works propose multi-scale DNN models to handle perspective/scale variations across scenes \cite{sindagi2017generating, Kang2018Crowd, Jiang2019Crowd, Liu2019Context, yang2020reverse, Jiang_2020_CVPR, Ma2019BayesianLF}.
\cite{kang2017incorporating} proposed an adaptive convolution neural network (ACNN), which utilizes the context (camera height and angle) as an auxiliary input to make the model  adaptive to different perspectives. 
\cite{shi2019revisiting} integrated the perspective information to provide additional knowledge of a person's scale change in an image.
\cite{Wang2019Learning} utilized a large synthetic single-view crowd counting dataset to train counting models, and used a CycleGAN transfer model to apply these models to real-world datasets, improving performance.

\textbf{Synthetic datasets.}
Synthetic data is an increasingly popular tool for training deep learning models for various computer vision tasks \cite{Nikolenko2019SyntheticDF},
such as \abc{single-view} crowd counting \cite{Wang2019Learning, COURTY2014161, cheung2016lcrowdv}, 
automatic driving \cite{li2019aads, saleh2018effective, hu2013batch}, 
image segmentation \cite{saleh2018effective},
and indoor navigation \cite{Savva2019HabitatAP, Savva2017MINOSMI, Song2017SemanticSC, Wu2018BuildingGA}.
Synthetic data is beneficial to real-world computer vision tasks when the real data is insufficient or difficult to acquire, or hard to annotate. 
To our knowledge, our generated dataset is the first large-scale synthetic dataset for the multi-view crowd counting problem.

\textbf{Cross-view cross-scene in other multi-camera tasks.}
Cross-view cross-scene is also an important issue for other multi-camera or cross-camera related vision tasks, such  as multi-view tracking/detection \cite{baque2017deep, chavdarova2018wildtrack}, 3D reconstruction \cite{Sitzmann2019SceneRN}, 3D human pose estimation \cite{iskakov2019learnable}, or person ReID \cite{zhong2018camera}.
%
%
\zq{To obtain cross-view or cross-scene generalization, these methods rely on large training datasets \cite{iskakov2019learnable}, image style transfer \cite{zhong2018camera} or adaptive view or scene modeling \cite{eslami2018neural, Sitzmann2019SceneRN}.}
\cite{iskakov2019learnable} presented two solutions for multi-view 3D human pose estimation based on learnable triangulation methods, by combining 3D information from multiple 2D views, and showed that the model trained on Human3.6m generalized to other 3D human pose datasets.
\cite{zhong2018camera} proposed to use CycleGAN \cite{CycleGAN2017} to smooth the camera style disparities.
\cite{Sitzmann2019SceneRN} proposed Scene Representation Networks, which map world coordinates to a feature representation of local scene properties, and generalize to other scenes by assuming 
the same class has common shape and appearance properties that are characterized by latent variables.
\cite{eslami2018neural} introduced the Generative Query Network to predict unobserved viewpoints, in which the viewpoint parameters are network inputs 
 and output is the 
view-adaptive feature representation.

 \zq{Similar to these approaches, we also need large-scale data for training the cross-view cross-scene MV counting models, and thus we generate a large synthetic dataset for the CVCS multi-view counting task. Furthermore, we propose an adaptive camera selection module to fuse multi-cameras, guided by the object-to-distance information. Unlike \cite{zhong2018camera}, we directly adapt the trained CVCS model to real-world datasets with unsupervised domain adaptation.}


\vspace{-0.1cm}
\section{CVCS Multi-View Counting}
\vspace{-0.2cm}


In this section, we describe the new cross-view cross-scene (CVCS) multi-view counting task.
We follow the camera settings of \cite{zhang2019wide, zhang2020_3d}, in which the input multi-camera views are synchronized and calibrated. However different from \cite{zhang2019wide, zhang2020_3d}, which assume a fixed number of fixed camera locations, CVCS assumes that the camera locations change, and the numbers and order of the cameras vary.
Most importantly, the model is trained and tested on distinct scenes and distinct camera layouts, so as to understand the cross-view and cross-scene generalization performance.


\subsection{CVCS multi-view counting model}

Our proposed CVCS multi-view counting model consists of 4 stages (see Fig.~\ref{fig:main_pipeline}), as follows.

{\em (1) Single-view feature extraction}: The first 7 layers of VGG-Net \cite{simonyan2014very, li2018csrnet} are used to extract the single-view features. To handle a variable number of views and input camera order, the feature extraction subnet is shared across all input camera views, which requires the feature extraction part to be general enough for different scenes and camera views. Thus, we choose VGG-Net as the feature extractor.

{\em (2) Single-view feature projection}: The extracted single-view features are projected to a common scene plane by a projection layer with variable camera parameters.
As in other multi-view methods \cite{zhang2019wide, zhang2020_3d, iskakov2019learnable, kar2017learning}, the projection is implemented with a spatial transformer net (STN) \cite{jaderberg2015spatial}. In contrast to \cite{zhang2019wide, zhang2020_3d}, which uses a fixed set of camera parameters in the projection, our model uses different camera parameters based on the current camera-views.

{\em (3) Multi-camera selection and fusion}: An adaptive CNN selects among the feature maps of the camera-views, based on an attention mechanism guided by the object-to-camera distance (more details in Sec. \ref{sec:camera_sel}).
The motivation of using a selection mechanism is to allow the network to learn,  at each location in the scene-level plane,  which camera should be more important 
during fusion.
A view-wise max-pooling layer fuses the projected multi-view features. Since the output size of the  max-pooling is fixed, the fusion stage is invariant to the number and order of the cameras.

{\em (4) Multi-view decoding}: 
the fused projected features are decoded to predict the scene-level density map.

The detailed layer settings of each module can be found in the supplemental.
The CVCS model is trained using the MSE loss between the predicted and ground-truth scene-level density map.
To make the CVCS model robust to non-correspondence errors arising from small camera calibration errors or erroneous features, we propose a noise-injection method that creates an extra camera view with noise (see Sec. \ref{sec:noise_view}).

%
%

A key difference of our CVCS model and the previous MV counting models \cite{zhang2019wide} is that our model is specifically designed to handle different scenes and camera views, through: 1) layers that are invariant to camera order and camera view, and 2) a multi-view fusion model based on camera layout geometry, which adapts to each camera layout.
Another key difference is  the proposed noise-injection method to regularize the our model during training.


\vspace{-0.1cm}
\subsection{Camera selection module} \label{sec:camera_sel}
\vspace{-0.2cm}

\begin{figure}[t]
\centering
   \includegraphics[width=0.9\linewidth]{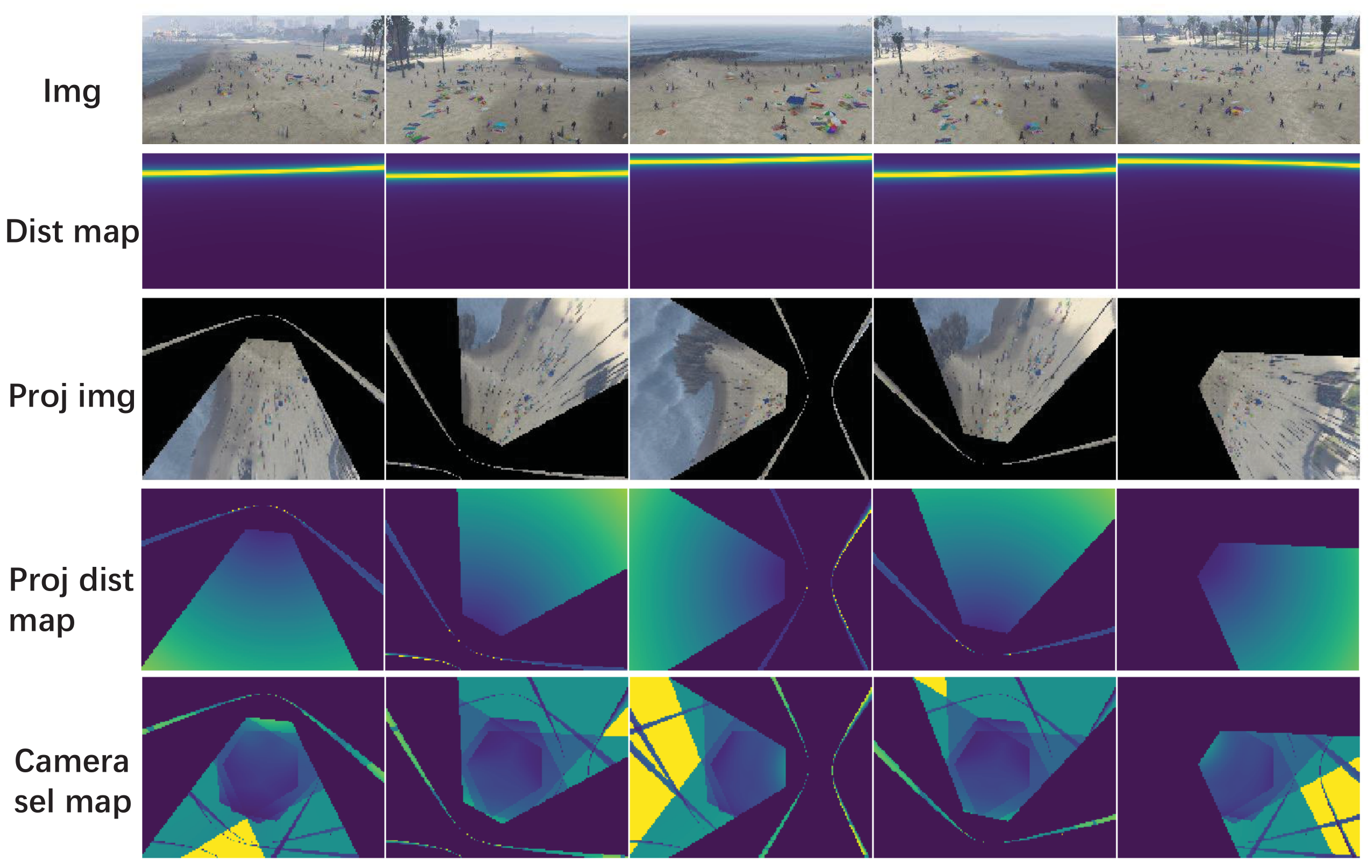}
   \vspace{-0.3cm}
   \caption{ The distance maps and camera selection maps.
   \zqq{Warmer colors indicate larger values. In the bottom row, yellow corresponds to 1.}
   }
      \vspace{-0.6cm}
\label{fig:noiseeg}
\end{figure}

\abc{Intuitively, when fusing multi-view features together to form the scene-plane feature map,
cameras that are closer to a particular location in the scene-plane should be favored over other cameras, since closer cameras have a clearer view of the location and likely yield more reliable features.}
%
Thus, we propose a camera selection module that selects and fuses cameras based on the object-to-camera distance.
The raw object-to-camera distance map is input into a subnet CNN, which maps the raw distance to a camera \aaabc{score}. 
Next, \aaabc{using the camera score,} a distance measure layer calculates the weights for each camera.

Formally, let $F$ and $P$ denote the feature extraction and projection layer. 
The $K$ input camera views are $\{V_k\}_{k=1}^K$,  their corresponding distance maps are $\{D_k\}_{k=1}^K$ (see Fig.~\ref{fig:noiseeg}) and extracted single-view features are $\{F_k\}_{k=1}^K$. The distance map of each camera view is computed 
(as in the side information used in \cite{zhang2019wide} for scale selection):
\begin{align}
  D(x, y) = \log(||R{\cal P}(x, y, h_{avg}) + T||_2),\label{eq:distance_map}
\end{align}
where $\cal P$, $R$ and $T$ are the camera-view to world projection function, the rotation matrix, and translation, respectively, and $h_{avg}$ is the average person height.

To perform camera selection, first the distance maps are passed through a shared CNN $M$, and then projected to the scene-plane, $M_k = P(M(D_k))$.
\zq{Next, the distance measure layer transforms $M_k$ into  weight for each camera:}
\begin{align}
  {\tilde{W}_k} =  1/\exp({(M_k-\hat{M})^2}),
  \label{eqn:Wk}
\end{align}
where $\hat{M}$ is the distance to the nearest camera for each pixel $(i,j)$ on the scene-plane,
\begin{align}
  \hat{M}(i,j) =  \min(M_1(i,j), \cdots, M_K(i,j)). \label{eq:min_distance_map}
\end{align}
In (\ref{eqn:Wk}), the nearest camera is assigned weight 1 and other cameras are assigned weights proportionally to the distance ratio, yielding the weight map.
\zq{Note that the pixels out of the camera views are masked out, and not involved in the camera selection map calculation.}
%
The weight maps are normalized across views,
  $W_k =  {\tilde{W}_k}/\sum^{K}_{n=1}{\tilde{W}_n}$,
%
\zq{
and for each view $k$, the weight map is element-wise multiplied with 
the corresponding projected camera-view feature map $P(F_k)$, yielding the attended projected feature map $W_k\otimes P(F_k)$.}


\begin{figure}[t]
\centering
   \includegraphics[width=0.9\linewidth]{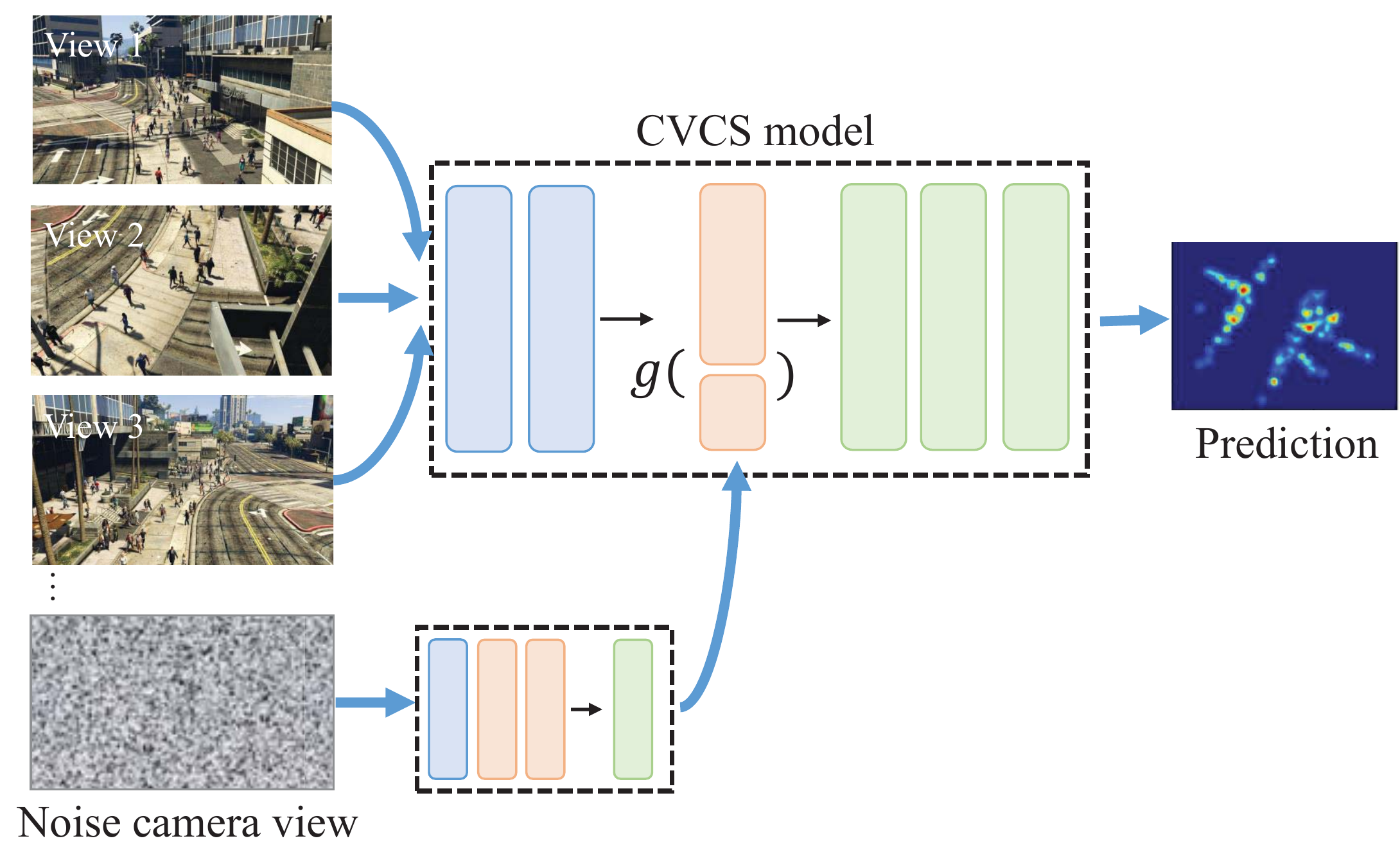}
   \vspace{-0.3cm}
   \caption{CVCS model with noise view regularization during training. The noise camera view can be added at different layers (see Tab.~\ref{tab:gs}). During testing, the noise view is removed.}
\label{fig:noise_view_module}
   \vspace{-0.4cm}
\end{figure}

In summary, the camera selection module uses side information to dynamically select and fuse the camera views based on the geometry of the camera layout.
For comparison, a scale selection module is used in \cite{zhang2019wide}, where the object-to-distance map selects the appropriate feature scale in the camera-view to obtain scale consistency across views and \aaabc{within} images.  In contrast, our camera selection module picks the appropriate camera when performing feature fusion at the scene-level, in order to keep the highest fidelity features.
\abc{Adaptive camera selection is required for CVCS, since the camera layout changes. In contrast, for single-scene MV models, the camera selection is learned implicitly in the network parameters, which does not generalize to new camera layouts.}
\abc{We further compare these two selection modules in the ablation study.}


\subsection{Noise camera view regularization}\label{sec:noise_view}

\begin{figure}[t]
\centering
   \includegraphics[width=0.9\linewidth]{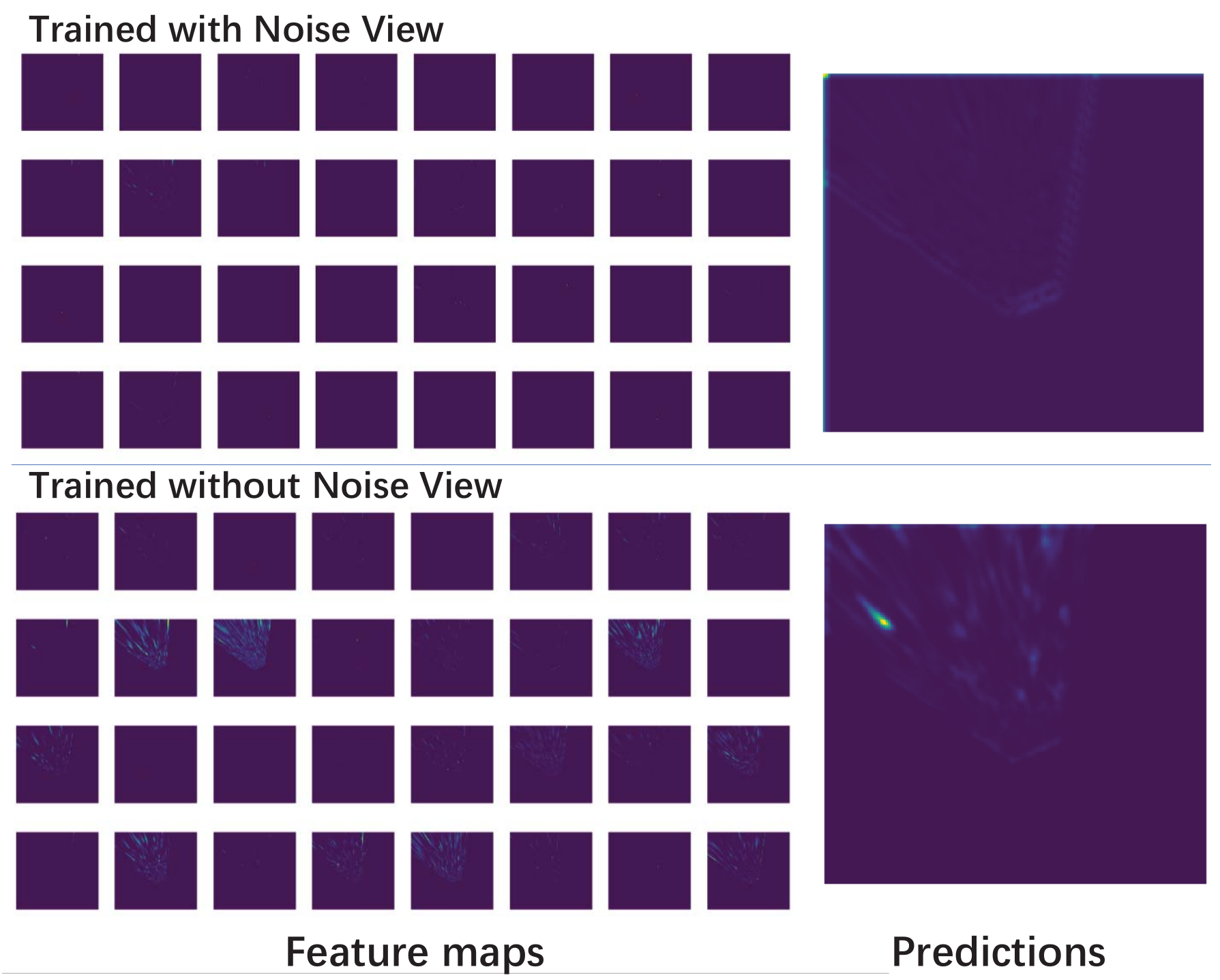}
      \vspace{-0.3cm}
   \caption{
   Robustness to noise:
   the feature maps and prediction for models trained with/without noise view regularization.
   The input is a Gaussian noise image.
   }
   \vspace{-0.4cm}
\label{fig:noise_view_feature}
\end{figure}

\abc{Assuming a correct camera projection and noiseless feature extraction process, the projected features should all align on the scene-level plane. However, in a real system, camera calibration errors, imperfect projection operators, and spurious noise in the feature maps cause {\em non-correspondence errors} in the projected feature maps. In single-scene MV counting, a {\em scene- and layout-dependent} CNN fusion module learns how to handle these non-correspondence errors. On the other hand, CVCS counting cannot use layout-dependent fusion modules, since the scene and camera layout change.}

To make the CVCS model robust to non-correspondence error, we propose a noise-based regularization scheme, where an extra camera view consisting of Gaussian noise $\epsilon = N(0,1)$ is
input in the model at training stage together with the real camera views (see Fig.~\ref{fig:noise_view_module}).
The camera geometry of the virtual noise camera is the same as one of the input cameras. Note that the noise camera view \aaabc{and associated layers} are  removed in the testing stage.



\aabc{Intuitively, the noise camera view simulates non-correspondence errors by randomly activating features in the map.} Training with the noise camera view guides the model to reduce the influence of \aaabc{this type of noise} on the final prediction,
preventing overfitting.
\aabc{An example is seen in Fig.~\ref{fig:noise_view_feature}, where the model learns to ignore the noise when trained with noise-view regularization.}
%
The regularization effect can also be explained from other aspects:

1) \textbf{Data augmentation}. Denote the whole CVCS model as $\cal M$. The model with noise camera view is changed from ${\cal M}(x_1, ..., x_{K})$ to ${\cal M}(x_1, ..., x_{K}, \epsilon)$, where $x_k$ is a real input camera view and $\epsilon$ is the random noise view. During training, the input camera views can be the same but noise view $\epsilon$ is random, which is a type of data augmentation.

2) \textbf{Noise injection}. The proposed noise camera view regularization approach can also be explained as a new noise injection function \cite{noh2017regularizing} for improving model generalization. The training of a noise-injected neural networks is equivalent to optimizing the lower bound of the marginal likelihood over noise $\epsilon$ \cite{noh2017regularizing}. The difference between additive Gaussian noise, dropout and the proposed noise camera view is the form of the noise injection function $g(x,\epsilon)$. 
Different noise injection functions $g$ can arise by changing the network layer to which the noise is injected (see Tab.~\ref{tab:gs}).
We compare these different noise injection functions in the ablation study (Sec.~\ref{text:ablation}).

\begin{table}
\centering
\footnotesize
\caption{Noise injection functions $g$ that vary on where the noise is added. $P$ is the projection layer, $F$ is the image feature extractor, and $H$ is a separate feature extractor for the noise.}
   \vspace{-0.3cm}

\begin{tabular}{@{}l@{\hspace{0.1cm}}l@{}}
$g(x,\epsilon)$ & Noise view added ... \\
\hline
   $\max(P(F(x)), \epsilon)$ & after projection \\
   $\max(P(F(x)), P(\epsilon))$ & before projection \\
   $\max(P(F(x)), P(F(\epsilon)))$ &  at input layer (same feature extractor) \\
   $\max(P(F(x)), P(H(\epsilon)))$ &  at input layer (different feature extractor) \\
   $\msum(P(F(x)), P(F(\epsilon)))$ &  at input layer, sum (same feature extractor) \\
   $\msum(P(F(x)), P(H(\epsilon)))$ &  at input layer, sum (different feature extractor) \\
   \hline
\end{tabular}

   \vspace{-0.3cm}
\label{tab:gs}
\end{table}

\vspace{-0.1cm}
\subsection{Unsupervised domain adaptation to real data}
\label{sec:domain_gap}
\vspace{-0.2cm}

\begin{figure}[t]
\centering
   \includegraphics[width=0.9\linewidth]{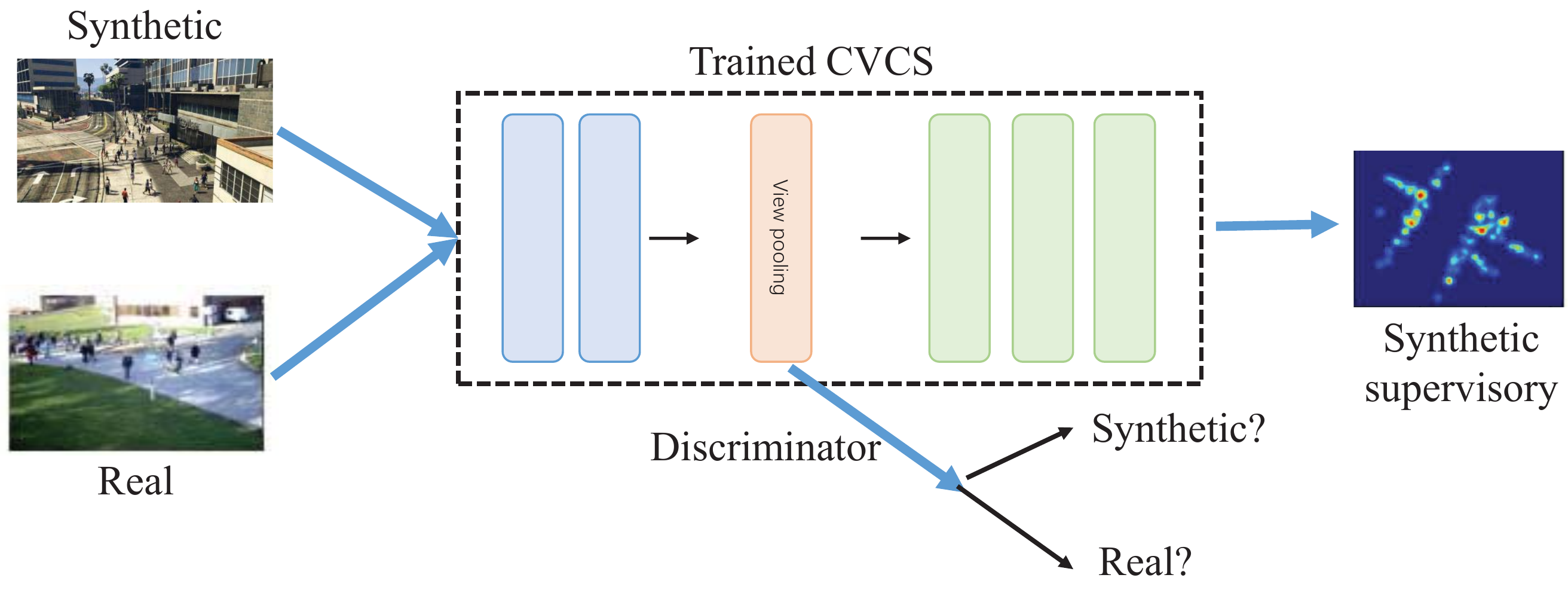}
      \vspace{-0.4cm}
   \caption{
   The unsupervised domain adaptation to real datasets. }
      \vspace{-0.5cm}
\label{fig:domain_gap}
\end{figure}

\abc{Our CVCS model is trained on a synthetic multi-view dataset (see Sec.~\ref{text:CVCSdataset}).
Directly applying the trained CVCS model to real multi-view datasets, such as PETS2009 and CityStreet, will be limited by the domain gap between synthetic and real scenes.}
\aabc{To reduce the domain gap, we apply unsupervised domain adaptation to fine-tune the trained CVCS model on each test scene, using only the test images (without the crowd labels).}
In particular, 
we add a feature discriminator to the \zq{trained} CVCS model to reduce the feature gap (see Fig.~\ref{fig:domain_gap}). 
The discriminator is inserted after the view-pooling layer in the CVCS model. During \aabc{the fine-tuning stage}, both the real and synthetic images are input to the model, then the real/synthetic features are fed to the discriminator to be classified. 
Only the synthetic features are sent to the multi-view decoder, since crowd labels for the real data are not available for training.
%
\zq{The fine-tuning loss function combines the synthetic counting loss and the discriminator loss}.
%
%
Our procedure of training on synthetic data and then testing on real data after unsupervised domain adaptation is more useful for practical applications, compared to previous state-of-the-arts \cite{zhang2019wide, zhang2020_3d}, which require training and testing on the same real scene with fixed cameras.

\vspace{-0.1cm}
\section{Synthetic CVCS Dataset}
\vspace{-0.2cm}
\label{text:CVCSdataset}

The proposed large synthetic multi-view crowd counting dataset is generated using GCC-CL \cite{Wang2019Learning}, which works as a plug-in for the game 
``Grand Theft Auto V''. The generating process consists of two parts: scene simulation and multi-view recording. First, crowd scenes are simulated, through the selection of the background selected, region of interest (ROI), weather condition, human models and postures, \emph{etc}. Next, cameras are placed at various locations to record the crowd scene from various perspectives. Birds-eye views are also collected for visualization.
Each person has a specific ID for mapping their coordinates in the world coordinate system and their locations in each camera-view image. The camera parameters, such as coordinates, deflection angles and fields-of-view, are also recorded.


\begin{table}[t]
  \centering
  \scriptsize
  \caption{Comparison of multi-view crowd datasets.
  }
	\vspace{-0.3cm}
  \begin{tabular}{@{}l@{\hspace{0.15cm}}@{\hspace{0.15cm}}c@{\hspace{0.15cm}}c@{\hspace{0.15cm}}c@{\hspace{0.15cm}}c@{\hspace{0.15cm}}c@{}}
  \hline
   Dataset     & Imgs. (train / test)   & Scenes & Counts & Views  & Image Res.   \\
  \hline
  PETS2009    & 1105 / 794 & 1      & 20-40  & 3      & 768$\times$576     \\ 
  DukeMTMC    & 700 / 289  & 1      & 10-30  & 4      & 1920$\times$1080 \\ 
  CityStreet & 300 / 200  & 1      & 70-150 & 3      &  2704$\times$1520 \\ 
  Synthetic (ours) & \textbf{200,000} / \textbf{80,000}  & \textbf{31}   & \textbf{90-180} & \textbf{60-120} & 1920$\times$1080 \\
  \hline
  \end{tabular}\label{tab:compare_dataset}
	\vspace{-0.4cm}
\end{table}



 In total, the whole synthetic MV counting dataset contains 31 scenes. For each scene, around 100 camera views are set for multi-view recording. The multi-view recording is performed 100 times with different crowd distributions in the scene, i.e., each scene contains 100 multi-view frames, with each frame comprising 60 to 120 camera-views. The image resolution is 1920$\times$1080. Compared with other MV counting datasets, like PETS2009 \cite{ferryman2009pets2009}, DukeMTMC \cite{ristani2016MTMC} and CityStreet \cite{zhang2019wide}, our proposed synthetic dataset contains more scenes, more camera views variations, and more total images (see Tab.~\ref{tab:compare_dataset}), which makes it more amenable for training and  validating CVCS multi-view counting.
Example images from the proposed synthetic multi-view counting dataset are shown in Fig.~\ref{fig:examples} \abc{and the Supplemental}.


\vspace{-0.1cm}
\section{Experiments}
\vspace{-0.2cm}

In this section, we conduct experiments on CVCS multi-view counting using our proposed model.

\subsection{Test datasets}\label{sec:dataset}

The real test datasets are PETS2009 \cite{ferryman2009pets2009}, DukeMTMC \cite{ristani2016MTMC} and CityStreet \cite{zhang2019wide}.
We use the same dataset settings as previous MV counting \cite{zhang2019wide}.
The dataset information is shown in 
Tab.~\ref{tab:compare_dataset}.
The input images are downsampled to 384$\times$288, 640$\times$360, and 676$\times$380, for PETS2009, DukeMTMC, and CityStreet, respectively.
The ground-plane scene-level density maps have resolutions of 152$\times$177, 160$\times$120, and 160$\times$192 for the three datasets.



\subsection{Experiment settings}

The synthetic dataset contains 31 scenes in total, of which 23 scenes are used for training and the remaining 8 scenes are used for testing. During training, we randomly select $K=5$ views  for $P=5$ times in each iteration per frame of each scene. For evaluation, we randomly select $K=5$ views for $P=21$ times ($V/K+1$, $V=100$ camera views) per frame of each scene, in order to test on more camera layouts. %
The input image resolution is 640$\times$360 (resized to $1/3$ of the original resolution).
Ground-plane patch-based training is used instead of complete image training, where 5 patches are extracted from the view-pooled features, corresponding to the 5 patches extracted from the ground-truth scene-level density maps. The patch size is 160$\times$180, and 1 pixel is equal to 0.5m in the real world. The ground-truth density maps are generated by convolving a  Gaussian kernel with the ground-truth person annotation dot map.
The learning rate is 1e-3, with learning decay 1e-4, and weight decay is 1e-4.
\aabc{The single-view feature extractor is pretrained via a single-view counting task on the synthetic dataset.}
Two evaluation metrics are used, the mean absolute error (MAE) and mean normalized absolute error (NAE) of the predicted counts on the test set.

\subsection{Experiment results}\label{sec:experiment}

\abc{We first report results on CVCS multi-view counting on the synthetic dataset, followed by the ablation study.  Finally, we present results on the real datasets.}

\begin{table}[]
  \centering
  \caption{Experiment results of our CVCS model with different  modules.}
   \vspace{-0.3cm}
\small
  \begin{tabular}{l|cc}
  \hline
   Model                 & MAE   & NAE    \\ \hline
  Backbone        & 14.13      & 0.115       \\ \hline
  Backbone+MVMS              & 9.30     & 0.080       \\ \hline
  Backbone+CamSel            & 8.63     & 0.074       \\
  Backbone+NoiseV            & 7.94     & 0.069      \\  
  CVCS (Backbone+CamSel+NoiseV)  & \textbf{7.22}     & \textbf{0.062}     \\ \hline
  \end{tabular}\label{tab:cvcs_results}
\end{table}

\begin{figure}[t]
\centering
   \includegraphics[width=\linewidth]{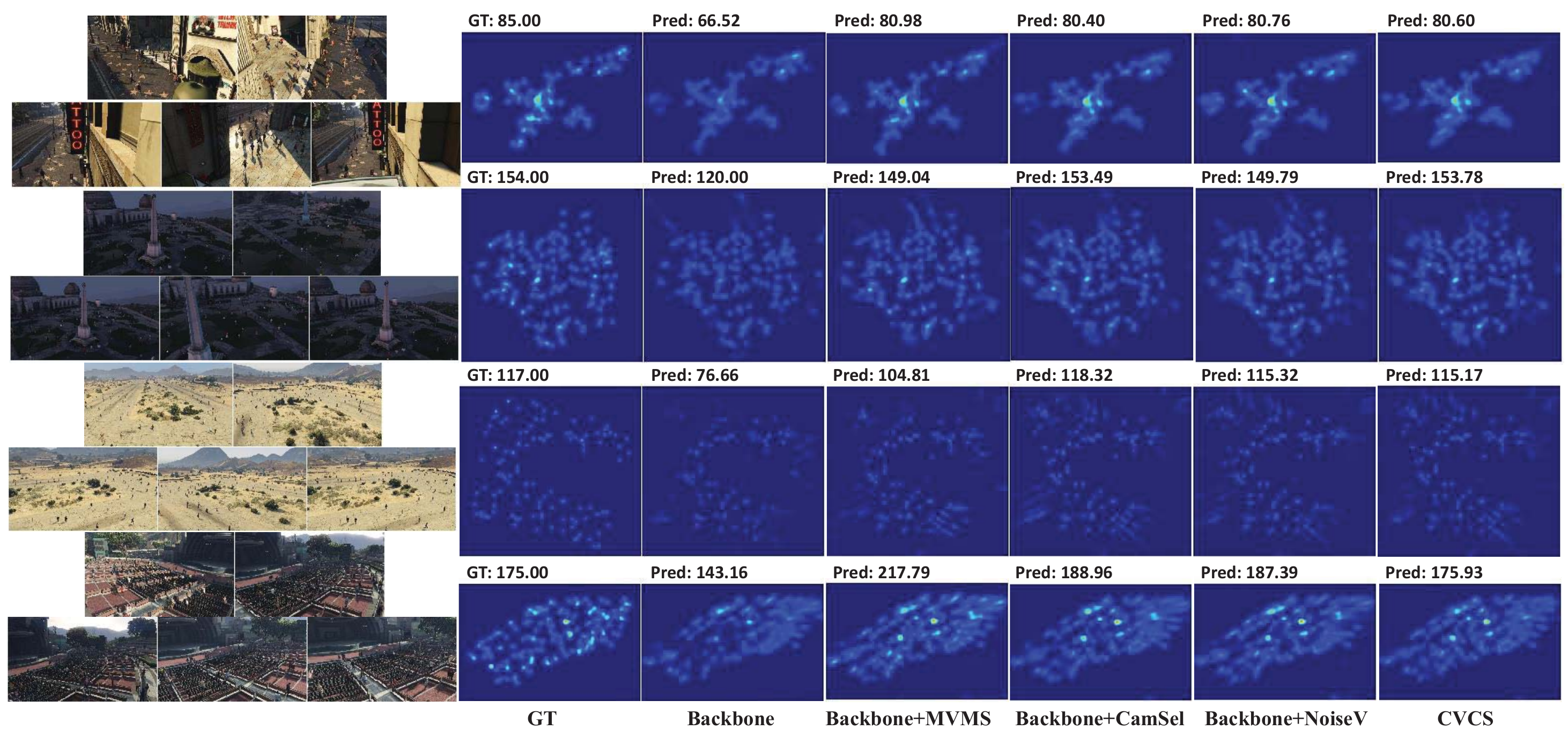}
   \vspace{-0.6cm}
   \caption{\zqq{The results of CVCS variations on synthetic datasets.
   Using camera selection  and/or noise-view regularization (CVCS, Backbone+CamSel, Backbone+NoiseV) are more accurate than the backbone or backbone with MVMS (Backbone+MVMS). See Supplemental for full-size figure.}
   \NOTE{this figure not referenced in text.  it has the same label as Fig 8.}}
   \vspace{-0.5cm}
\label{fig:syn_results}
\end{figure}

\vspace{-0.2cm}
\subsubsection{CVCS performance}
\vspace{-0.2cm}

We test 5 variations of our CVCS model on the synthetic dataset.
The first method is the backbone model of our CVCS model without the camera selection and noise camera view (denoted as ``Backbone'').
The second method is the backbone model with the multi-view multi-scale selection architecture from \cite{zhang2019wide} (denoted as ``Backbone+MVMS''), where a 3-scale pyramid is used in the feature extraction part and camera view distance maps are used to fuse the multi-scales before projection. \aaabc{The next 2 methods add either the camera selection module (Backbone+CamSel) or the noise camera view (Backbone+NoiseV) to the backbone model.  Finally, adding both modules to the backbone yields our full model (CVCS).}


The results on the synthetic dataset are shown in Table~\ref{tab:cvcs_results} and Fig.~\ref{fig:syn_results}.
Using the proposed camera selection module or noise view regularization yields large reduction of errors over the Backbone, while their combination further reduces the errors, showing that the two modules can work together to improve CVCS performance in terms of different aspects.
The MVMS architecture also reduces the error when used with the Backbone, as the  multi-scale selection can promote scale consistency of the features across camera views. However, the error reduction using MVMS is not as large as using CamSel and NoiseV. The main reason is that the camera selection and noise view regularization modules are designed for better cross-view cross-scene performance.


\begin{table}[]
  \centering
  \caption{Ablation study on the camera selection module.}
  	\vspace{-0.3cm}
\footnotesize
  \begin{tabular}{l|cc}
  \hline
   Model                  & MAE   & NAE    \\ \hline
Backbone        & 14.13      & 0.115       \\ \hline
  +CamSel (no conv)              & 10.77     & 0.089        \\ 
  +CamSel ($1\!\times\!1$ conv)   & 8.63     & 0.074       \\
  +CamSel (3 conv)    & 8.15     & 0.069      \\  \hline
  \end{tabular}\label{tab:cam_results}
  	\vspace{-0.3cm}
\end{table}

\vspace{-0.2cm}
\subsubsection{Ablation study}
\vspace{-0.2cm}
\label{text:ablation}

In the ablation study we consider various permutations of our CVCS model.

\textbf{Camera selection module.}
We consider 3 settings of the CNN mapping in the camera selection module:
no convolution layers, i.e., passing the distance map directly to the distance measure layer (denoted as ``no conv''),
a
1$\times$1 conv layer,
 and a 3-layer CNN. The results are shown Table \ref{tab:cam_results}. Compared to the Backbone model, all the camera selection settings reduce the counting error; CNN achieves the best error reduction due to its flexibility in mapping the distance information to a suitable weight.
%
%
%
%

\begin{table}[]
  \centering
  \caption{Ablation study of the noise view regularization.}  	
  \vspace{-0.3cm}

\footnotesize
  \begin{tabular}{l@{\hspace{0.2cm}}l|c|c}
  \hline
   Model & Type                     & MAE   & NAE              \\ \hline
  \multicolumn{2}{l|}{Backbone}            & 14.13      & 0.115       \\ \hline
  \multicolumn{2}{l|}{+Dropout}               & 13.16      & 0.111       \\ \hline

  +NoiseV &(A) $\max(x,\epsilon)$              & 19.91    & 0.163        \\ \hline


  +NoiseV & (B) $\max(P(F(x)),\epsilon)$              & 9.64     & 0.084        \\
  +NoiseV & (C) $\max(P(F(x)),P(\epsilon))$             & 9.42     & 0.079   \\
  +NoiseV & (D) $\max(P(F(x)),P(F(\epsilon)))$           & \textbf{7.94}     & \textbf{0.069}        \\
  +NoiseV & (E) $\msum(P(F(x)),P(F(\epsilon)))$          & 8.57     & 0.076        \\
  +NoiseV & (F) $\max(P(F(x)),P(H(\epsilon)))$           & 8.48     & 0.074        \\
  +NoiseV & (G) $\msum(P(F(x)),P(H(\epsilon)))$          & 8.54     & 0.076        \\ \hline
  \end{tabular}\label{tab:noise_results}
  \vspace{-0.3cm}
\end{table}

\textbf{Noise view regularization.}
We next experiment with different types of the noise view regularization added to the Backbone, as shown Table \ref{tab:noise_results}.
%
Generally, adding noise-based regularization can improve the performance, except when directly corrupting the input image (NoiseV, Type A).
%
\abc{The best regularization occurs when the noise is passed through a feature extractor and projection (Types D-G). These noise injection functions better simulate the noisy feature extraction process and how the noise is projected into the scene-level feature map.}
Dropout regularization is also tested with the Backbone, where 2 dropout layers are added after the 2nd and 4th layers of the feature extractor.
Dropout can also improve the performance slightly, but not as much as the noise camera view.


%
%

\begin{table}[]
  \centering
  \footnotesize
  \caption{Ablation study combining the camera selection module and noise view regularization. Noise types are defined in Table \ref{tab:noise_results}.}
  \vspace{-0.3cm}

  \begin{tabular}{l|c|c}
  \hline
   Model     & MAE   & NAE    \\ \hline
  Backbone        & 14.13      & 0.115       \\ \hline
  +CamSel (1$\times$1 conv)+NoiseV (Type D) 
  & \textbf{7.22}     & \textbf{0.062}       \\
  +CamSel (1$\times$1 conv)+NoiseV (Type E) 
  & 9.98       & 0.087       \\
  +CamSel (1$\times$1 conv)+NoiseV (Type F) 
   & 8.34     & 0.074       \\
  +CamSel (1$\times$1 conv)+NoiseV (Type G) 
  & 8.32       & 0.074       \\\hline

  +CamSel (3 conv)+NoiseV (Type D) 
  & 7.96    & 0.069       \\
  +CamSel (3 conv)+NoiseV (Type E) 
  & 8.22     & 0.072      \\
  +CamSel (3 conv)+NoiseV (Type F) 
   & 7.56     & 0.066       \\
  +CamSel (3 conv)+NoiseV (Type G) 
  & 7.44      & 0.065       \\\hline

  \end{tabular}\label{tab:combine_results}
  \vspace{-0.2cm}

\end{table}

\textbf{Combining camera selection and noise view regularization.}
From the previous experiments, we find that the camera selection and noise view regularization can both improve the performance of Backbone model. Since the two modules address different aspects, we combine the two modules together to further improve the performance. Specifically, the camera selection modules are combined with different noise view regularization methods and the results are shown in Table \ref{tab:combine_results}. The best combination uses $1\times1$ conv in the camera selection module and noise injection in the input view (Type D), $\max(P(F(x)),P(F(\epsilon)))$.
Generally, the various combinations perform better than the Backbone and Backbone+MVMS models.

\begin{table}[]
  \centering
  \caption{Ablation study on different numbers of input camera views.}
  	\vspace{-0.3cm}
\footnotesize
  \begin{tabular}{@{}c|cc|cc@{}}
  \hline
       & \multicolumn{2}{c|}{Backbone}   & \multicolumn{2}{c}{CVCS}  \\ \hline
  No. Views & MAE & NAE & MAE & NAE \\  \hline
  3     & 14.28 & 0.130      &  7.24 & 0.071         \\ 
  5     & 14.13 & 0.115     & 7.22 &0.062      \\ 
  7     & 14.35 & 0.113   & 7.07 &0.058     \\
  9     & 14.56 & 0.112   & 7.04 &0.056   \\
  11    & 15.15 & 0.115   & 7.00 &0.055   \\ \hline
  \end{tabular}\label{tab:cam_num_results}
	\vspace{-0.3cm}

\end{table}

\begin{figure*}[t]
\centering
   \includegraphics[width=0.7\linewidth]{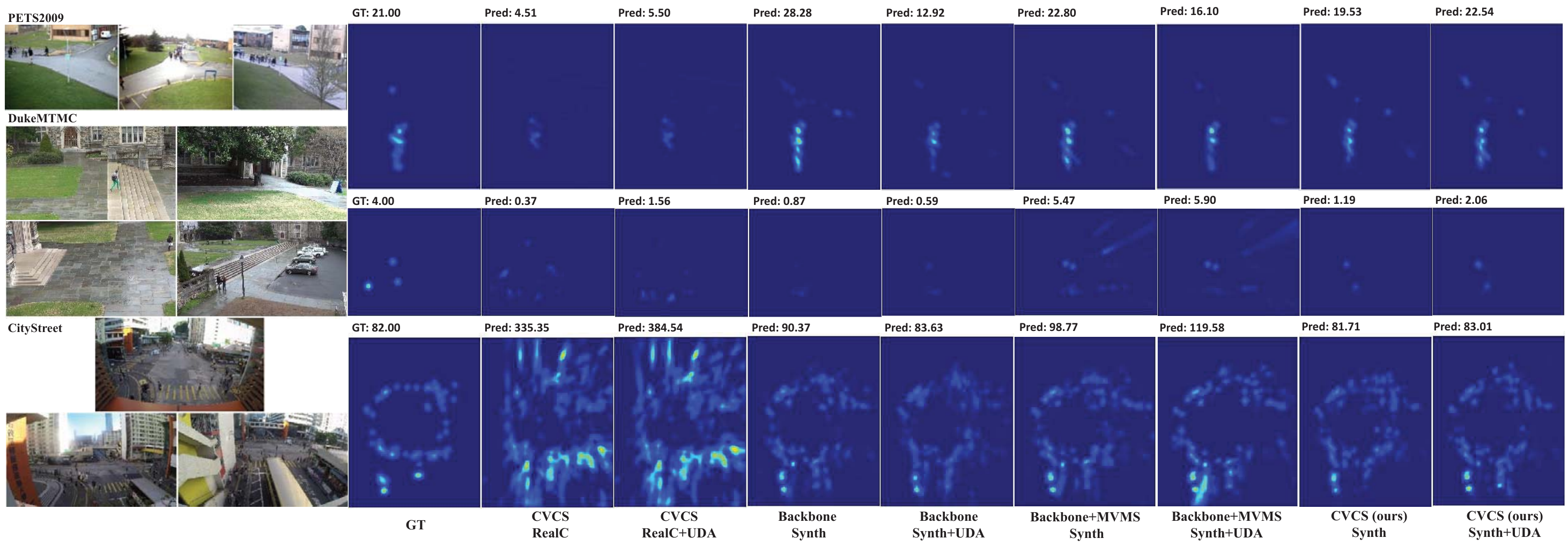}
    \vspace{-0.3cm}
   \caption{\zqq{The cross-view cross-scene results on real datasets. Our CVCS model trained on the synthetic data shows better performance than CVCS trained on real data. Applying unsupervised domain adaptation (UDA) to our CVCS improves the performance. See Supp.~for full-sized figure.}}
\label{fig:real_results}
 \vspace{-0.4cm}
\end{figure*}

\textbf{Variable camera number.}
Our CVCS model is specifically designed to handle any number of camera views at test time.
To show the influence of number of camera views, we test CVCS with different number of input cameras. 
Note that the models are trained with 5 input camera views, and tested on different number of views.
The results are presented in in Table \ref{tab:cam_num_results}.
The performance with different numbers of cameras is stable for both the Backbone and the full model CVCS. 
Increasing the camera views from 5 to 11, the full CVCS achieves slightly lower error because the extra cameras provide more information in some regions that were poorly covered with only 5-cameras.



%

\begin{table}[]
  \centering
  \footnotesize
  \caption{Results testing on real datasets. Different training schemes are used: ``RealS" means training and testing on the same real scene (single-scene MV); ``RealC'' means cross-scene training on 2 real scenes and testing on the other; ``Synth'' means cross-scene training on synthetic dataset; ``+UDA" adds unsupervised domain adaptation.
    }
    \vspace{-0.3cm}
  \begin{tabular}{@{}l@{\hspace{0.13cm}}l@{\hspace{0.05cm}}|@{\hspace{0.05cm}}r@{\hspace{0.05cm}}r@{\hspace{0.13cm}}r@{\hspace{0.05cm}}r@{\hspace{0.13cm}}r@{\hspace{0.05cm}}r@{}}
\hline
  & & \multicolumn{6}{c}{Test Scene} \\
  &   &  \multicolumn{2}{c}{PETS2009}    &    \multicolumn{2}{c}{\hspace{-0.25cm}DukeMTMC}        &  \multicolumn{2}{c}{CityStreet}        \\
Model &  Training & MAE & NAE &  MAE & NAE & MAE & NAE \\
  \hline
  Dmap\_wtd \cite{Ryan2014Scene}   & RealS 		& 7.51    & 0.261      & 2.12   & 0.255      & 11.10  & 0.121      \\
  Dect+ReID \cite{zhang2019wide}   & RealS			& 9.41 & 0.289  & 2.20 & 0.342 & 27.60 & 0.385   \\
  LateFusion \cite{zhang2019wide}  & RealS 		 & 3.92  &0.138  & 1.27 &0.198 & 8.12  &0.097  \\
  EarlyFusion \cite{zhang2019wide} & RealS	 & 5.43 &0.199 & 1.25&0.220 & 8.10 & 0.096\\
  MVMS \cite{zhang2019wide} & RealS		     & 3.49 &0.124  & \textbf{1.03} & \textbf{0.170}  & 8.01 & 0.096 \\
  3D \cite{zhang2020_3d} & RealS		     & \textbf{3.15} &\textbf{0.113 }& 1.37  &0.244 & \textbf{7.54} & \textbf{0.091} \\
   \hline
   \hline
  CVCS   & RealC	          & 23.34 & 0.729 & 5.28 & 0.623 & 215.23 & 2.700 \\
  CVCS   & RealC+UDA	      & 20.11 & 0.636 & 5.34 & 0.628  & 249.25 & 3.110  \\
  \hline
  Backbone & 	Synth	 & 8.05 & 0.257  & 4.19 & 0.913 & 11.57 & 0.156  \\
  Backbone+MVMS  & 	Synth	    & 6.03 & 0.191  & 3.07 & 0.553 & 14.02 & 0.194  \\
  CVCS (ours) & 	Synth	 & 5.33 & 0.174  & 2.85 & 0.546              & 11.09 & 0.124  \\
  Backbone & Synth+UDA    & 5.91 & 0.200  & 3.11 & 0.551 & 10.09 & 0.117   \\
  Backbone+MVMS  & Synth+UDA    & 5.28 & 0.175  & 3.00 & 0.585 & 12.05 & 0.157   \\
  CVCS (ours) & Synth+UDA	 &  {\bf 5.17} & {\bf 0.165}  & {\bf 2.83} & {\bf 0.525} & {\bf 9.58} & {\bf 0.117}   \\
  \hline
  \end{tabular}\label{tab:real_results}
  \vspace{-0.5cm}
\end{table}

\vspace{-0.2cm}
\subsubsection{Cross-view cross-scene on real data}
\vspace{-0.2cm}

We apply the trained CVCS model to real multi-view counting datasets.
Unlike previous state-of-the-arts \cite{zhang2019wide,zhang2020_3d} that train and test on the same single real scene with fixed camera views, we train on synthetic scenes and test on real scenes (cross-scene setting) with different camera views (cross-view). We allow unsupervised domain adaptation, which uses the images of the test scene to fine-tune the feature extractor, and does not use crowd labels.
We believe this testing paradigm is more practical, since annotations of the target scenes are not required.

We test three groups of methods, with different training setups.  The first group is trained and tested on \emph{single} real scenes (denoted as ``RealS''), and include 2 traditional multi-view counting methods, Dmap\_wtd \cite{Ryan2014Scene, zhang2019wide} and Dect+ReID \cite{ren2015faster, liao2015person, zhang2019wide}, 3 DNN-based fusion methods (EarlyFusion, LateFusion, and MVMS) \cite{zhang2019wide}, and 3D counting \cite{zhang2020_3d}.
The second group uses cross-scene training on the real datasets (denoted as ``RealC''), where our CVCS model (Backbone+CamSel+NoiseV) is trained on 2 real scenes, and tested on the remaining scene.
The third group uses cross-scene training on our synthetic dataset and directly tests on the real data (denoted as ``Synth''). \aaabc{We test 3 models: Backbone, Backbone+MVMS, and our CVCS model.}
We also add unsupervised domain adaptation using images from the test scene (``+UDA''). Note that no crowd labels are used from the test scene for UDA.


The test performances of these methods on the 3 real datasets
are shown in Table \ref{tab:real_results}.
Directly testing our synthetically trained CVCS on real scenes achieves promising performance,
which is better or competitive to the 2 traditional multi-view counting methods, \zq{Dmap\_wtd and Dect+ReID}.
Furthermore, adding unsupervised domain adaptation (``Synthetic+UDA'') effectively reduces the domain gap between synthetic and real data, yielding lower errors compared to without UDA.
%
%
\aaabc{Our CVCS also outperforms Backbone and Backbone+MVMS, for both Synth training and Synth+UDA training, which shows the advantage of our cross-scene cross-view specific modules, camera selection and noise-view regularization.}

Training using only the real scenes in a cross-scene manner (RealC) yields very large errors, compared to training with the synthetic data (Synth), \abc{showing that the model overfits when there are too few training scenes, and that there is significant benefit in training on more scenes/views even if they are synthetic.}
\abc{Finally, our CVCS model 
using UDA is slightly worse than the MVMS model \cite{zhang2019wide} trained and tested on the same-scene (MAE increases by 1-2), which shows the promise of our approach and the CVCS paradigm.} Example visualizations of the results are presented in Fig.~\ref{fig:real_results}.

\vspace{-0.15cm}
\section{Conclusion}
\vspace{-0.05cm}
\aabc{In this paper, we propose the task of cross-view cross-scene (CVCS) multi-view counting, where models are trained to generalize to different scenes and camera layouts.
We propose a CVCS multi-view DNN with a camera selection and fusion module and noise-view regularization, to adapt the network to different camera layouts and to learn to ignore non-correspondence errors.
We collect a large-scale synthetic dataset with large numbers of camera views and scenes for training and evaluating the CVCS multi-view counting.
%
Furthermore, we show that the synthetically-trained CVCS model can be applied to real scenes via unsupervised domain adaptation, which only uses images from the test scene.
Overall, our work 
advances research on multi-view crowd counting from the single-scene fixed-camera setting to cross-view cross-scene setting, which is more practical for deployment.
}

{\small
\textbf{Acknowledgments}. This work was supported by a grant from the Research Grants Council of the Hong Kong Special Administrative Region, China (Project No. CityU 11212518).
}

{\small
\bibliographystyle{ieee_fullname}
\bibliography{egbib}
}

\end{document}